# A MULTIMODAL APPROACH FOR ENDOSCOPIC VCE IMAGE CLASSIFICATION USING *BiomedCLIP-PubMedBERT*

A PREPRINT

**Dr. Nagarajan Ganapathy***
Department of Biomedical Engineering
Indian Institute of Technology Hyderabad
Sangareddy, Hyderabad, India
gnagarajan@bme.iith.ac.in

**Podakanti Satyajith Chary**
Department of Biomedical Engineering
MedInfoLab, IIT Hyderabad
Sangareddy, Hyderabad, India
satyajithpodakanti@gmail.com

**Teja Venkata Ramana Kumar Pithani**
Department of Biomedical Engineering
Indian Institute of Technology University
Sangareddy, Hyderabad, India
pithani.tejavenkataramanakumar@gmail.com

**Pavan Kavati**
Department of Biomedical Engineering
Indian Institute of Technology University
Sangareddy, Hyderabad, India
bm23resch01001@iith.ac.in

**Arun Kumar S**
Department of Biomedical Engineering
Indian Institute of Technology University
Sangareddy, Hyderabad, India
bm24resch12001@iith.ac.in

October 25, 2024

## ABSTRACT

This Paper presents an advanced approach for fine-tuning BiomedCLIP-PubMedBERT, a multimodal model, to classify abnormalities in Video Capsule Endoscopy (VCE) frames, aiming to enhance diagnostic efficiency in gastrointestinal healthcare. By integrating the PubMedBERT language model with a Vision Transformer (ViT) to process endoscopic images, our method categorizes images into ten specific classes: angioectasia, bleeding, erosion, erythema, foreign body, lymphangiectasia, polyp, ulcer, worms, and normal. Our workflow incorporates image preprocessing and fine-tunes the BiomedCLIP model to generate high-quality embeddings for both visual and textual inputs, aligning them through similarity scoring for classification. Performance metrics, including classification accuracy, recall, and F1 score, indicate the model's strong ability to accurately identify abnormalities in endoscopic frames, showing promise for practical use in clinical diagnostics.

We are proud to share that our approach earned **2nd position in the Capsule Vision Challenge 2024**, reflecting the robustness and applicability of our methodology in addressing real-world challenges in endoscopic diagnostics.

## 1 Introduction

Video Capsule Endoscopy (VCE) is a widely utilized, non-invasive technique in gastrointestinal diagnostics, capturing thousands of images as it moves through the digestive tract. This method provides clinicians with valuable insights into gastrointestinal abnormalities, but the sheer volume of VCE frames poses a significant challenge for manual interpretation. The process is time-intensive, subject to human error, and requires specialized expertise, making it less efficient for clinical use at scale. Consequently, artificial intelligence (AI) methods have emerged as promising tools to automate abnormality detection, streamline diagnostic workflows, and reduce clinician workload.

* https://github.com/Satyajithchary/MedInfoLab_Capsule_Vision_2024_Challenge

The accurate classification of gastrointestinal abnormalities within VCE images, however, presents unique challenges. These include handling high intra-class variability, dealing with imbalanced datasets, and ensuring that models generalize well across different patient populations and imaging conditions. Previous approaches have primarily relied on supervised learning models that require large, annotated datasets, which are often difficult to obtain in medical imaging due to privacy concerns and the specialized knowledge required for labeling.

To address these challenges, this study employs a fine-tuning approach using BiomedCLIP—a multimodal vision-language model that combines the PubMedBERT language model with a Vision Transformer (ViT). BiomedCLIP leverages the strengths of PubMedBERT's domain-specific language processing and ViT's ability to capture spatial features within images, enabling a powerful classification system for endoscopic images. By generating embeddings for both visual and text data, the model aligns these embeddings via similarity scoring, enabling the classification of VCE frames into ten abnormality categories: angioectasia, bleeding, erosion, erythema, foreign body, lymphangiectasia, polyp, ulcer, worms, and normal.

This study aims to demonstrate the efficacy of fine-tuning BiomedCLIP on VCE images for accurate abnormality classification. We evaluate the model's performance using key metrics such as accuracy, precision, recall, and F1 score. The findings contribute to the growing body of research on AI-driven diagnostics in gastrointestinal healthcare and underscore the potential of vision-language models to improve diagnostic efficiency and accuracy in clinical settings.

## 2 Dataset

The dataset used in this study was provided by the **Capsule Vision 2024 Challenge: Multi-Class Abnormality Classification for Video Capsule Endoscopy[2]**, focusing on multi-class abnormality classification in Video Capsule Endoscopy (VCE) frames. This dataset includes labeled images across ten gastrointestinal categories: angioectasia, bleeding, erosion, erythema, foreign body, lymphangiectasia, polyp, ulcer, worms, and normal, providing a comprehensive foundation for fine-tuning AI models to automatically classify these abnormalities. The dataset was curated from both publicly available and proprietary sources to represent diverse imaging conditions, enhancing the model's generalizability.

### 2.1 Dataset Composition

The dataset was developed using three publicly available VCE datasets—SEE-AI, KID, and Kvasir-Capsule, as well as one private dataset from AIIMS. The training dataset consists of 37,607 frames, while the validation dataset contains 16,132 frames, both mapped to the ten abnormality classes: angioectasia, bleeding, erosion, erythema, foreign body, lymphangiectasia, polyp, ulcer, worms, and normal.

| Source | Training Frames | Validation Frames |
|---|---|---|
| KID | 376 | 165 |
| Kvasir | 26,511 | 11,581 |
| SEE-AI | 9,092 | 4,291 |
| AIIMS | 224 | 97 |
| **Total** | **37,607** | **16,132** |

Table 1: Dataset Composition

The dataset is organized into training and validation directories, with each class (e.g., Angioectasia, Bleeding) stored in a corresponding subfolder. The images are named and stored according to their respective abnormality class. Each set is accompanied by a metadata file (training-data.xlsx and validation-data.xlsx) containing the image paths and corresponding labels.

### 2.2 Data Organization and Metadata

Each abnormality class is represented as a subfolder within the training and validation directories, enabling systematic organization of the image data. For instance, images classified as "Angioectasia" are stored in a dedicated folder named after the class, making it easier for the model to associate images with corresponding labels. Additionally, metadata files (training-data.xlsx and validation-data.xlsx) accompany the dataset, listing each image's path and its label, which ensures smooth data retrieval during model training and validation.

### 2.3 Data Augmentation and Preprocessing

To address class imbalances and enhance model robustness, various data augmentation techniques were applied, including rotation, flipping, and cropping. Each image was resized to 224x224 pixels, matching the input size requirements of the Vision Transformer (ViT) model. Furthermore, we utilized the preprocess function from the openclip package to transform the images into tensor format, standardizing pixel values and ensuring compatibility with the BiomedCLIP architecture.

## 3 Methodology

This section describes the steps involved in developing and implementing a fine-tuned BiomedCLIP-PubMedBERT model to classify abnormalities in VCE images into ten categories. The methodology encompasses dataset preparation, model architecture, training process, and evaluation metrics. Each subsection provides a comprehensive view of the data flow and architecture used to achieve efficient classification in endoscopic frames.

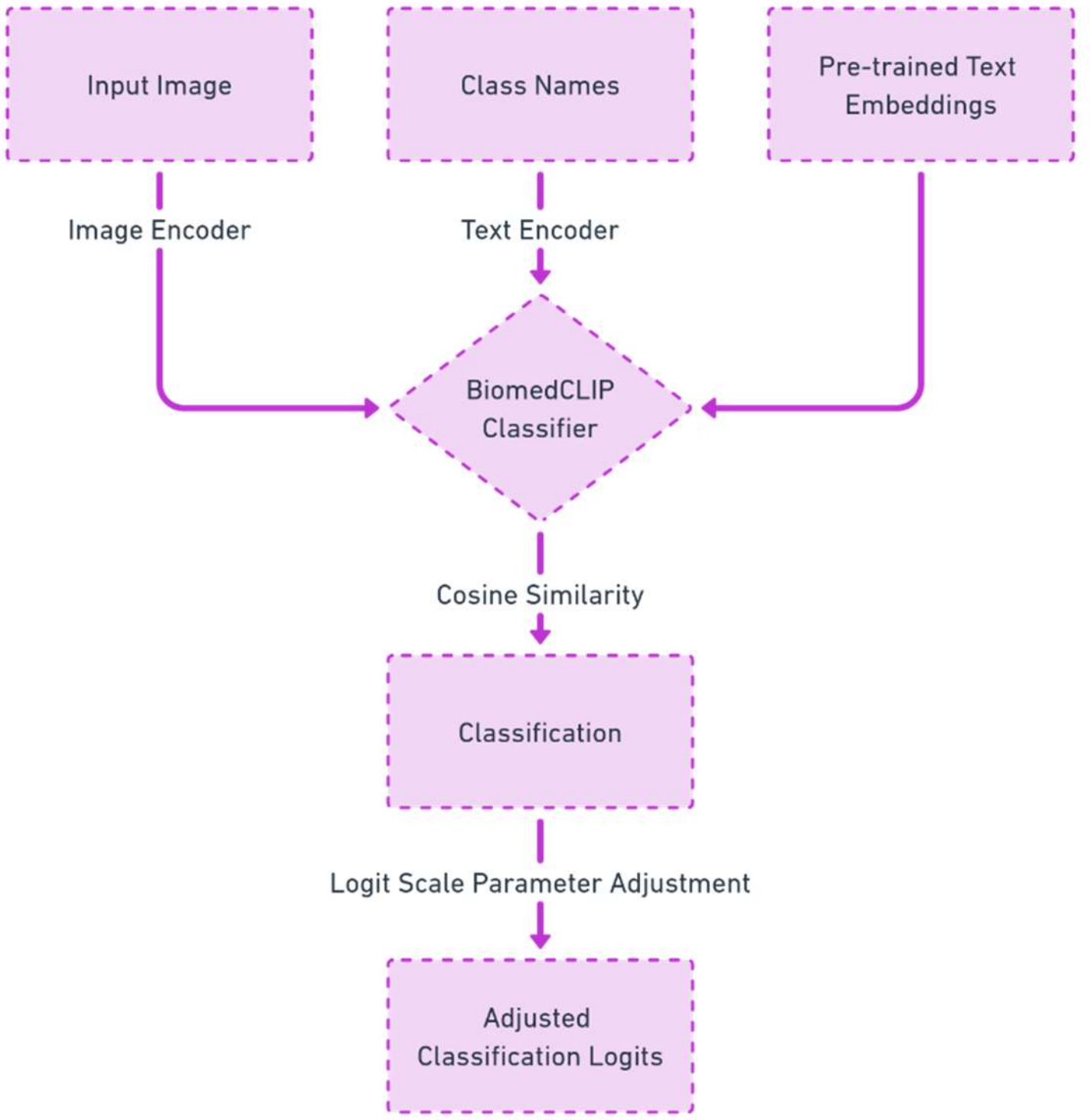

Figure 1: Pipeline of Project

### 3.1 Dataset Preparation

The dataset is structured into two parts: a training set and a validation set. Each set contains subfolders corresponding to the ten categories of abnormalities:

- **Training Set:** Contains labeled images within subfolders named after each of the ten categories.
- **Validation Set:** Contains labeled images used for internal validation.

Each image in the dataset corresponds to one of the ten abnormality classes. Additionally, an Excel file, training-data.xlsx and validation-data.xlsx, contains metadata about each image. The goal is to develop a model that can classify unseen test images into one of these ten categories.

The following steps were used to preprocess the data set:

- **Data loading:** The images were loaded from the dataset folders using Python's os and PIL.Image libraries. A loop iterates over each class folder to build a list of image paths and corresponding labels.
- **Data Augmentation and Re-Sizing:** Since medical imaging datasets are often imbalanced and small, augmentation techniques such as rotation, flipping, and cropping were employed to increase dataset variability. The images were resized to 224x224 pixels to match the input size of the transformer-based architecture of the model, which helps to reduce computational cost.
- **Preprocessing:** The preprocess function, provided by the *openclip* package, was used to transform the images into a tensor format compatible with the BiomedCLIP model. This function normalizes the pixel values and ensures that the input is in the correct shape for the model.

## 3.2 Model Architecture

The architecture of our model is based on BiomedCLIP, a pre-trained vision-language model designed specifically for medical image classification tasks. For the Capsule Vision 2024 Challenge, we employed BiomedCLIP-PubMedBERT integrated with a Vision Transformer (ViT) for the task of abnormality classification in Video Capsule Endoscopy (VCE) images. Below, we provide a detailed explanation of each component of the model, along with the data flow, technical specifications, and how these elements work together to address the multi-class classification task.

### 3.2.1 Model Overview and Flow

The model architecture is composed of two key components:

- **A Vision Transformer (ViT)**, which processes the endoscopic images and extracts image features.
- **PubMedBERT,** part of the BiomedCLIP framework, which generates embeddings from the text (i.e., abnormality class labels).

In the abnormality classification pipeline, these components work together as follows:

1. **Input Stage:**
    - Image Input: VCE images from the dataset are preprocessed into a format compatible with the Vision Transformer. Each image is resized to 224x224 pixels, which is standard for ViT models.
    - Text Input: The text-based class labels (e.g., "Angioectasia," "Bleeding," etc.) are tokenized and transformed into embeddings using PubMedBERT.
2. **Feature Extraction:**
    - The Vision Transformer (ViT) processes the image and extracts a sequence of features. These features capture the important spatial details necessary for classifying the image.
    - Simultaneously, the PubMedBERT model processes the text labels, converting them into semantic embeddings. This step links the textual meaning of each abnormality class to its corresponding medical imagery.
3. **Image-Text Matching:**
    - BiomedCLIP then computes the similarity between the visual and textual embeddings. The model generates a similarity score for each class label, indicating how closely the extracted image features match the text embeddings.
4. **Classification Output:**

- Finally, a softmax layer is applied to the similarity scores to generate probabilities for each of the ten abnormality classes. The class with the highest probability is selected as the predicted abnormality for that image.

This fine-tuning approach allows the model to classify abnormalities without needing explicit labels for every possible abnormality in the training phase, a critical advantage for medical imaging tasks where fully labeled datasets are scarce.

### 3.2.2 Detailed Breakdown of Model Components

Here, we explain each part of the architecture in more technical detail, specifically tailored for the task of abnormality classification in VCE images.

1. **Vision Transformer (ViT)**
    - **Input:** The ViT model accepts an input image of size 224x224x3 (height, width, and RGB channels).
    - **Patch Embedding:** The image is divided into 16x16 non-overlapping patches. Each patch is flattened into a single 256-dimensional vector, then linearly projected to a 768-dimensional embedding. Given the 224x224 input, this results in 196 patches (14x14 grid), each represented as a 768-dimensional feature vector.
    - **Position Embeddings:** To retain spatial information, the model adds position embeddings to each patch embedding, maintaining information about each patch's location in the image, which is crucial for capturing spatial relationships between abnormality features.
    - **Transformer Encoder:** The patch embeddings are processed through multiple transformer layers. Each layer consists of multi-head self-attention mechanisms and feed-forward networks, which allow the ViT to capture long-range dependencies across patches and understand global image structures.
    - **Output:** The ViT produces a sequence of 196 feature vectors, each of 768 dimensions, representing the encoded visual information across the entire image. A specific [CLS] token at the beginning of this sequence is used as a summary of the image's features, which will be aligned with the textual embedding in the cross-modal matching stage.

2. **BiomedCLIP-PubMedBERT for Text Embeddings**
    - **Input:** PubMedBERT receives textual descriptions of each abnormality class (e.g., "This is an endoscopic image of Angioectasia"), designed to generate contextualized embeddings reflective of medical terminology and concepts.
    - **Tokenization:** Text input is tokenized with a vocabulary specialized for biomedical terms, allowing recognition of domain-specific language. Each class description is tokenized up to a maximum sequence length of 256 tokens, with token embeddings initialized based on PubMedBERT's pre-trained weights.
    - **Contextual Embedding Layers:** PubMedBERT uses a stack of 12 transformer encoder layers, each containing multi-head attention and feed-forward networks. These layers capture the semantic relationships between tokens, adapting BERT's general-purpose embeddings to interpret biomedical context effectively.
    - **Pooling and Output Representation:** The [CLS] token, a special token added to the beginning of each text input, is used as a summary embedding for the entire sequence. After processing through all transformer layers, this [CLS] embedding becomes the final 768-dimensional text embedding representing each class label, capturing the biomedical semantics specific to the abnormality.

3. **Cross-modal Similarity Matching**
    - **Embedding Alignment in a Shared Space:** After generating image embeddings from ViT and text embeddings from PubMedBERT, BiomedCLIP projects both embeddings into a shared multimodal feature space. This alignment allows direct comparison between visual features and semantic representations of abnormality classes.
    - **Similarity Calculation Using Dot Product:** The model computes similarity scores between each image embedding and the ten class text embeddings using a dot product. The dot product quantifies the alignment between the visual features of the image and the semantic meaning of each class description, resulting in a logit score for each class.

- **Logit Scale Adjustment:** A learnable scalar parameter, called the logit scale, is applied to the similarity scores. This scaling factor adjusts the logits, calibrating the model's confidence in its predictions and helping prevent overly confident misclassifications.

4. **Softmax Classification**
   - **Softmax Function for Probability Distribution:** After similarity scores are computed for all classes, a softmax layer converts these scores into a probability distribution over the ten classes. Each probability represents the model's confidence that the input image belongs to a particular class.
   - **Class Prediction:** The model selects the class with the highest probability as the final prediction for the image. This results in a 10-dimensional output vector, where each entry corresponds to the probability of the image belonging to each abnormality class.

### 3.2.3 Data Flow and Technical Specifications

- **Image Input Size:** 224x224x3 (for the Vision Transformer). Text Input Size: Maximum of 256 tokens (for PubMedBERT).

- **Vision Transformer Output:** A sequence of image features of size 196x768.

- **PubMedBERT Output:** A semantic embedding of size 768 for each class label.

- **Similarity Calculation:** Dot product between image and text embeddings, outputting a similarity score for each class.

- **Final Output:** A 10-dimensional probability vector representing the likelihood of each class, followed by the predicted class label.

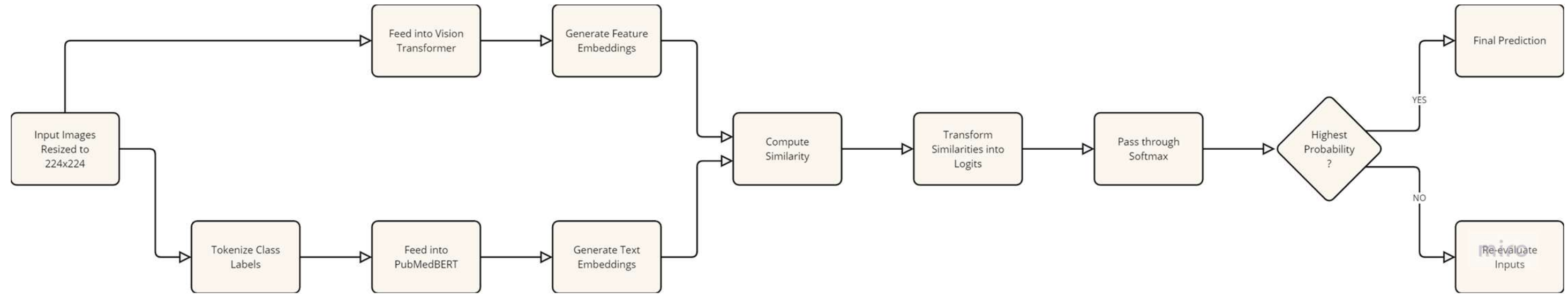

Figure 2: Block Diagram of the Developed Pipeline.

### 3.2.4 Strengths of the Model for Abnormality Classification

- **Fine-Tuning Capability:** The model's ability to classify abnormalities without requiring large amounts of labeled training data makes it highly efficient in medical scenarios where labeled datasets are limited.

- **Vision-Text Alignment:** By aligning image features with textual medical descriptions, BiomedCLIP leverages the power of large-scale pre-training, enabling it to generalize across unseen abnormalities.

- **Scalability:** The use of transformers for both vision and language tasks makes the model highly scalable. It can be fine-tuned or adapted for other medical imaging classification tasks with minimal changes.

## 4 Training Process

Although BiomedCLIP is a pre-trained model, additional fine-tuning was performed using the labeled training dataset. The steps involved in training are as follows:

- **Batch Processing:** Images were processed in mini-batches of 32, allowing efficient utilization of GPU memory. For each batch, VCE frames were fed through the Vision Transformer, producing a set of image embeddings. Simultaneously, text embeddings for the ten classes were loaded and matched with each image in the batch.
- **Similarity Computation:** For each image in the batch, similarity scores were calculated between the image embedding and each of the ten class embeddings. This process produced a set of logits, representing the model's confidence in assigning the image to each class.
- **Backpropagation:** Using the computed cross-entropy loss, gradients were backpropagated through the model's layers. The optimizer then adjusted the weights of the ViT and PubMedBERT components, fine-tuning them to improve class alignment.
- **Performance Tracking:** During training, accuracy and loss metrics were monitored at the end of each epoch. By evaluating these metrics on both training and validation sets, we tracked the model's convergence and prevented overfitting.
- **Epochs and Early Stopping:** Training was conducted over 30 epochs with early stopping criteria. If validation loss plateaued over multiple epochs, training was halted to avoid overfitting while preserving model generalizability.

## 5 Results

This section details the results obtained from the Fine-Tuning for classification model using the BiomedCLIP-PubMedBERT architecture on the Capsule Vision 2024 Challenge dataset. We evaluate the model's performance using multiple metrics, including accuracy, precision, recall, F1-score, and area under the ROC curve (AUC). The classification results are generated for each of the ten abnormality classes.

### 5.1 Model Performance Metrics

The classification performance of the model is evaluated on the validation dataset, which contains labeled images from the same ten abnormality classes as the training set. Below are the key evaluation metrics used to assess the model:

- **Accuracy:** Overall classification accuracy, i.e., the percentage of correct predictions.
- **Precision:** Measures the proportion of true positive predictions out of all positive predictions for each class.
- **Recall (Sensitivity):** Indicates the model's ability to correctly identify true positives out of all actual positive instances.
- **F1 Score:** The harmonic mean of precision and recall, providing a balanced measure when both metrics are considered important.
- **ROC AUC:** The area under the ROC curve, indicating how well the model distinguishes between classes across various decision thresholds.

### 5.2 Training and Validation Performance Analysis

### Training Results:

- **Epoch 1:** Train Loss: **0.3927** | Train Accuracy: **87.67%**
- **Epoch 2:** Train Loss: **0.2061** | Train Accuracy: **93.26%**
- **Epoch 3:** Train Loss: **0.1108** | Train Accuracy: **96.17%**

**Validation Results:**

- **Epoch 1:** Val Loss: **0.2621** | Val Accuracy: **91.46%**
- **Epoch 2:** Val Loss: **0.2369** | Val Accuracy: **92.36%**
- **Epoch 3:** Val Loss: **0.1822** | Val Accuracy: **94.04%**

The final model achieved an impressive **Training Accuracy of 97.75%** and **Validation Accuracy of 94.04%** by the third epoch, indicating a robust training process with minimal overfitting.

The model demonstrates strong performance in distinguishing between positive and negative cases across classes, as evidenced by the high AUC values observed in the **ROC curves for both training and validation sets**.

● The **training ROC curves**2 for each of the ten classes (Angioectasia, Bleeding, Erosion, Erythema, Foreign Body, Lymphangiectasia, Normal, Polyp, Ulcer, and Worms) show **AUC values of 1.00**, indicating perfect discriminatory ability during training.

● The validation ROC curves3 show slight performance degradation compared to the training curves but still maintain **high AUC values, ranging from 0.99 to 1.00**.

● This minor performance difference between training and validation suggests **Good Generalization Ability** of the model.

### 5.3 Class-Specific Performance

The model's performance varied across classes, with strong results in certain areas and challenges in others:

**Training Set Metrics:**

- **Balanced Accuracy:** 0.9388
- **Mean AUC:** 0.9990
- **Mean Average Precision (MAP):** 0.9774
- **Mean F1 Score:** 0.9389

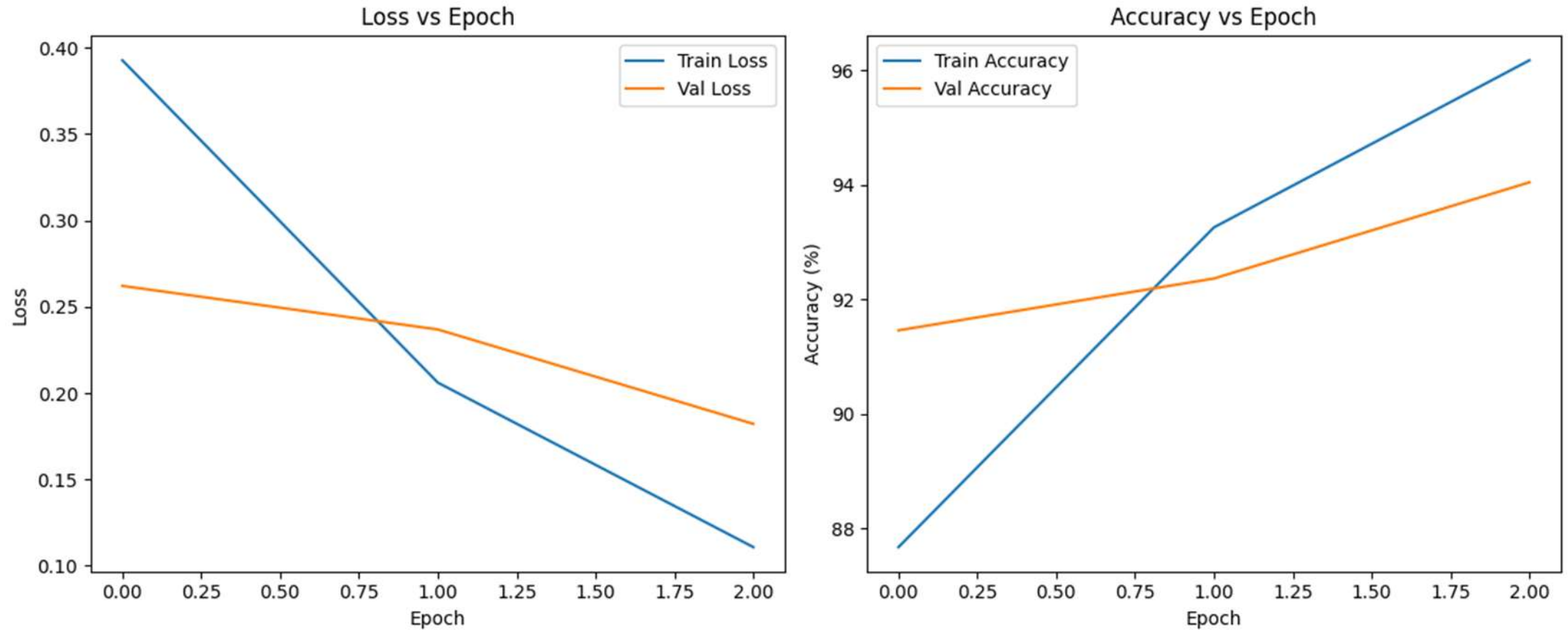

**Validation Set Metrics:**

- **Balanced Accuracy:** 0.8464
- **Mean AUC:** 0.9940
- **Mean Average Precision (MAP):** 0.9093
- **Mean F1 Score:** 0.8539

**Challenging Classes:**

- **Polyp Detection:** Lower AUC-ROC at 0.17, indicating difficulties in accurate detection.
- **Worms:** AUC-ROC at 0.23, reflecting limited detection capability.
- **Bleeding:** Moderate performance with an AUC-ROC of 0.25.

## 5.4 ROC Curves for Training and Validation

To assess the model's ability to distinguish between classes at various thresholds, ROC curves for both the training and validation sets were analyzed:

- **Training ROC Curve**:

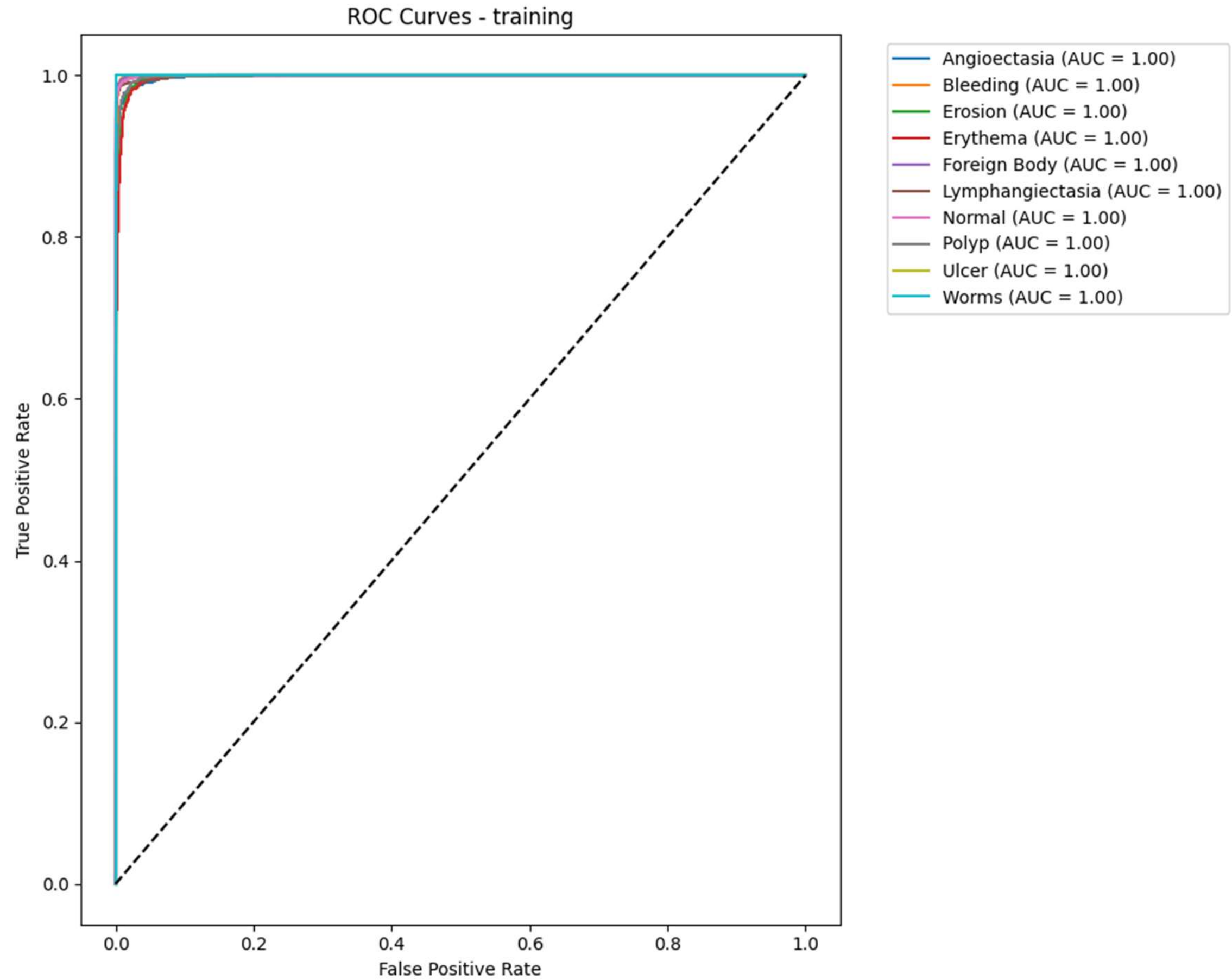

-

- **Validation ROC Curve**:

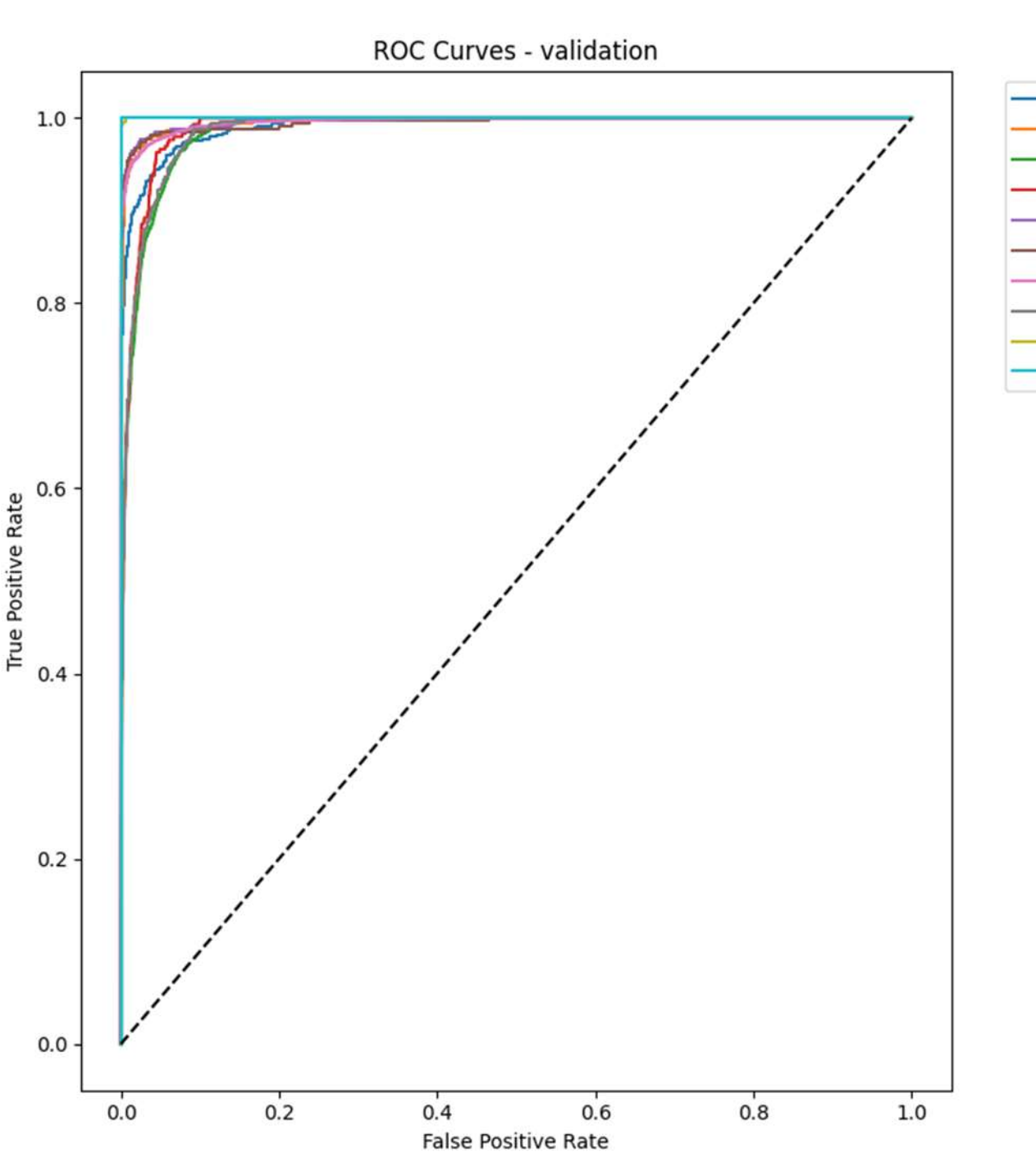

The ROC curves **demonstrate high AUC values**, showing that the **model effectively distinguishes between positive and negative cases across classes,** with minor performance degradation from training to validation, suggesting good generalization.

## 5.5 Confusion Matrices

The confusion matrices for training and validation further illustrate the model's performance by visualizing misclassifications:

- **Training Confusion Matrix**:

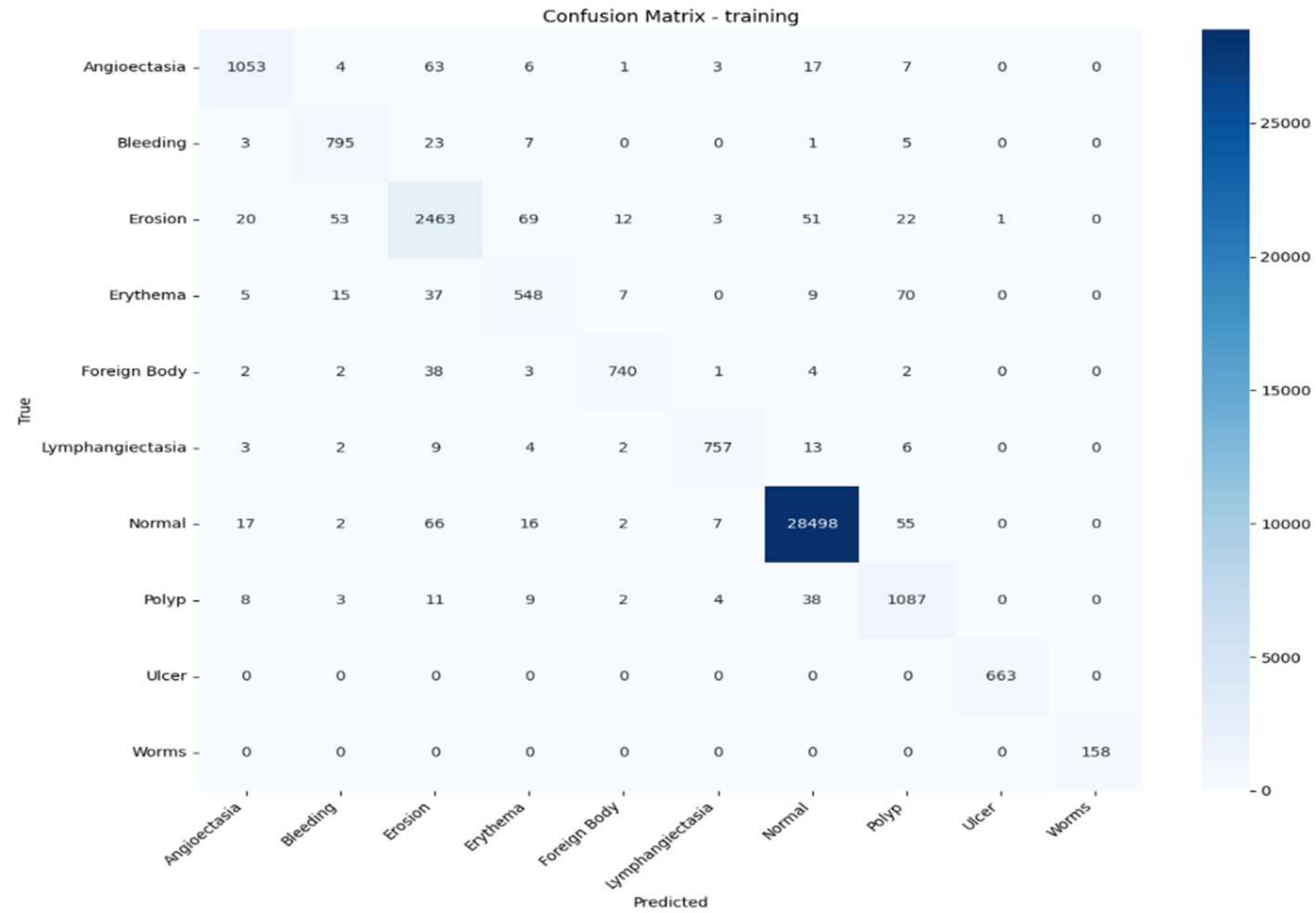

- **Validation Confusion Matrix**:

.

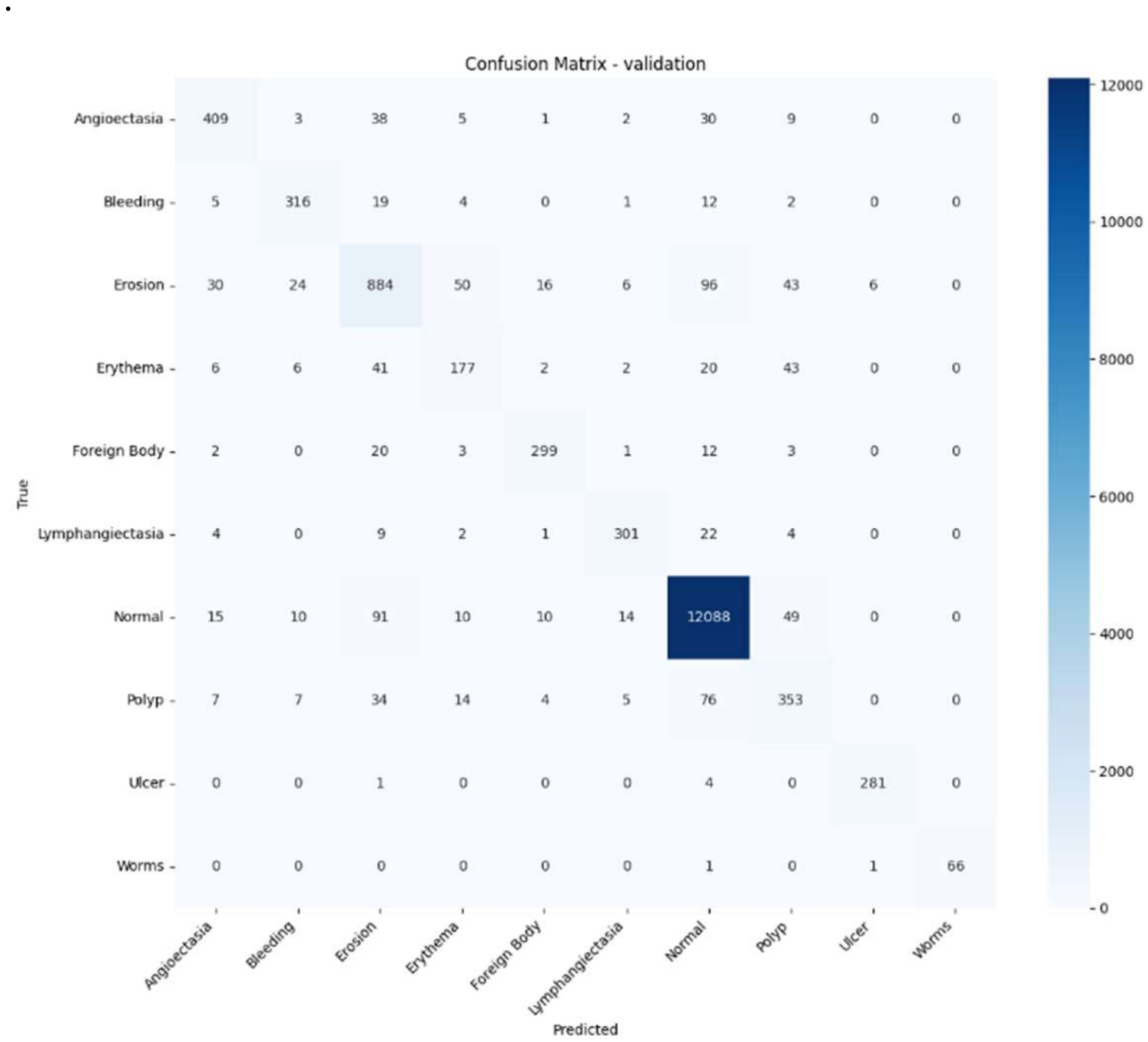

These matrices highlight areas where the model excelled (e.g., "Foreign Body") and struggled (e.g., "Erosion" vs. "Ulcer"), revealing class-specific performance variations.

- **Confusion matrices** provide a visual representation of the model's performance by highlighting correct predictions along the diagonal and misclassifications as off-diagonal entries.
- **Training confusion matrix** shows that the model correctly classified a significant number of instances for the **"Normal" class (28498) and "Erosion" class (2463).**
- However, there are also some misclassifications, particularly between the "Erosion" and "Ulcer" classes.
- Similarly, the **validation confusion matrix** shows that the model performed well in classifying "Normal" instances (12088) and "Erosion" (884).
- The validation confusion matrix also shows some misclassifications, particularly between "Erosion" and "Ulcer."

## 5.6 Per-Class Metrics

To gain more insight into class-specific performance, AUC, average precision, and F1 scores for each class were calculated on both the training and validation sets:

These visualizations indicate high performance for most classes, with slight drops in challenging classes such as "Erosion" and "Ulcer" on validation data.

- **Training Per-Class Metrics**:

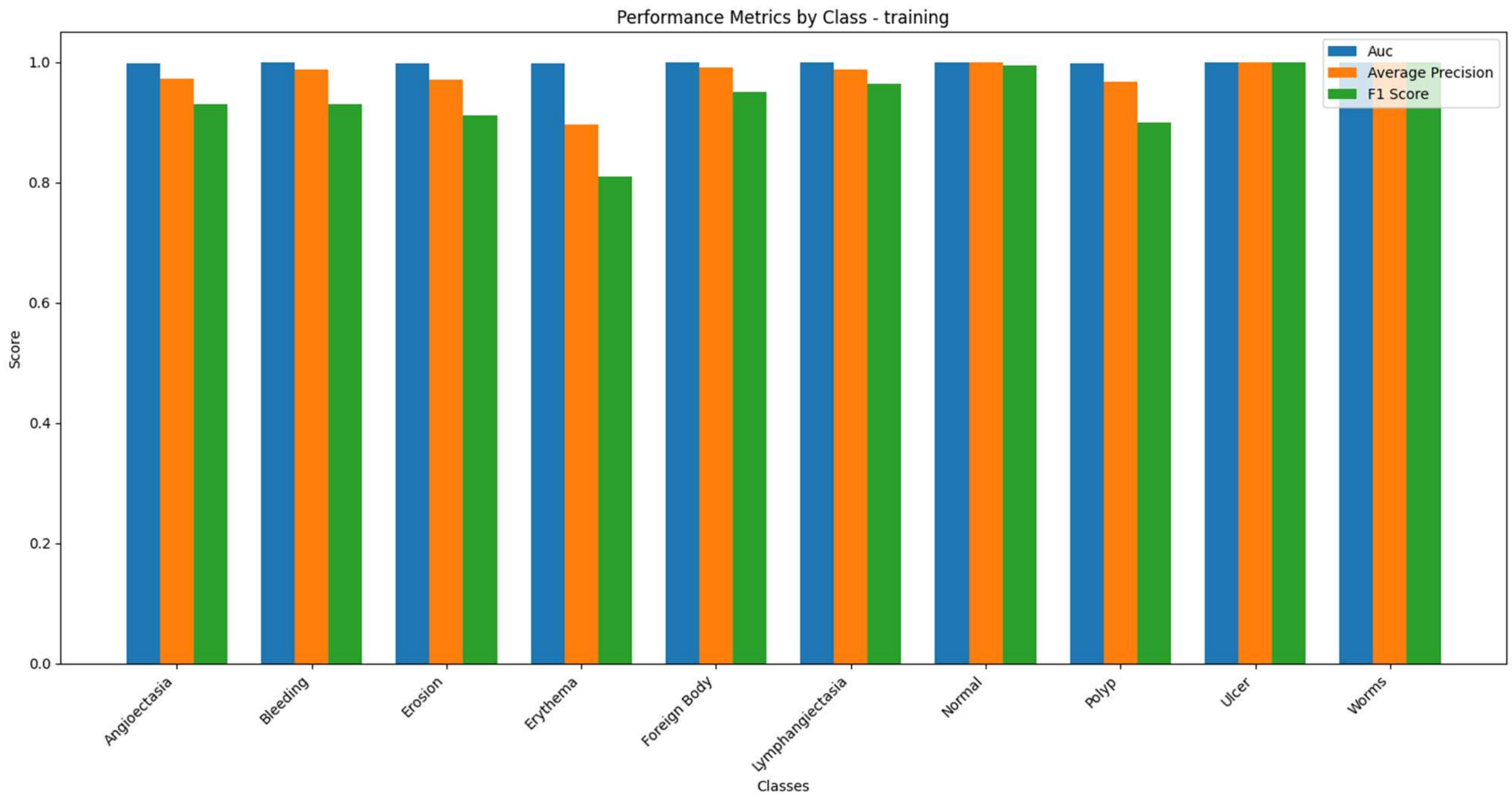

- **Validation Per-Class Metrics**:

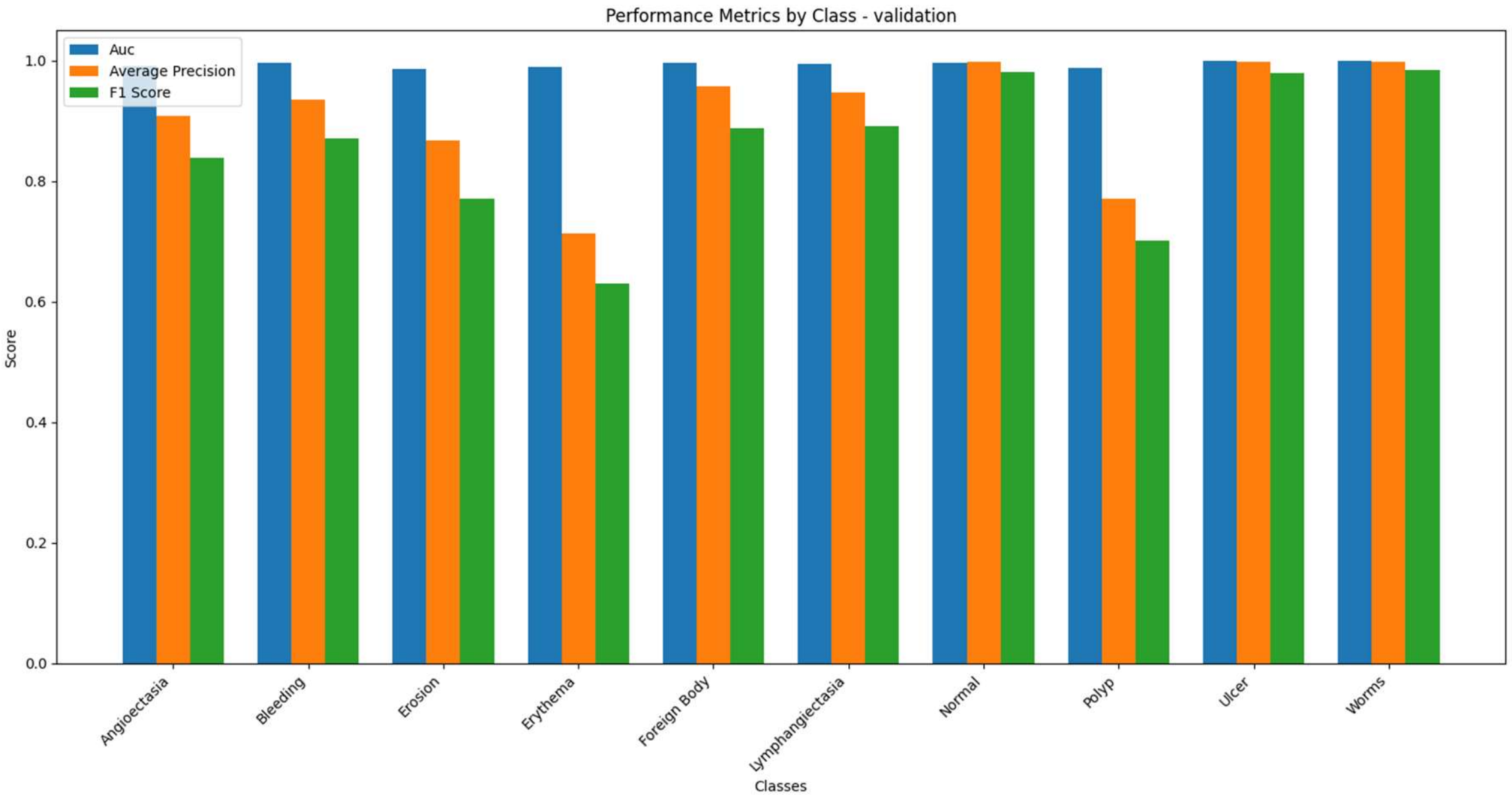

**Class-Specific Training AUC-ROC and F1 Scores:**

- **Angioectasia:** AUC-ROC: 0.9984 | F1 Score: 0.9298
- **Bleeding:** AUC-ROC: 0.9997 | F1 Score: 0.9298
- **Erosion:** AUC-ROC: 0.9978 | F1 Score: 0.9115
- **Erythema:** AUC-ROC: 0.9972 | F1 Score: 0.8101

- **Foreign Body:** AUC-ROC: 0.9998 | F1 Score: 0.9612
- **Lymphangiectasia:** AUC-ROC: 0.9992| F1 Score: 0.9637
- **Normal:** AUC-ROC: 0.9996 | F1 Score: 0.9948
- **Polyp:** AUC-ROC: 0.9986 | F1 Score: 0.8998
- **Ulcer:** AUC-ROC: 1.0000 | F1 Score: 0.9992
- **Worms:** AUC-ROC: 1.0000 | F1 Score: 1.0000

- Examining **per-class metrics** like AUC, average precision, and F1 scores provides a more granular understanding of the model's performance for each class.

- For instance, "Foreign Body," "Lymphangiectasia," "Normal," "Ulcer," and "Worms" achieve **high AUC-ROC and F1 scores in both training and validation** settings.

- Classes like "Erosion" and "Erythema" exhibit slightly **lower performance on validation data compared to training**, indicating potential challenges in accurately classifying these specific classes.

### 5.7 Precision-Recall Curve for Validation

The precision-recall curve for the validation set provides additional information on the trade-offs between precision and recall across different thresholds:

- **Training Precision-Recall Curve**:

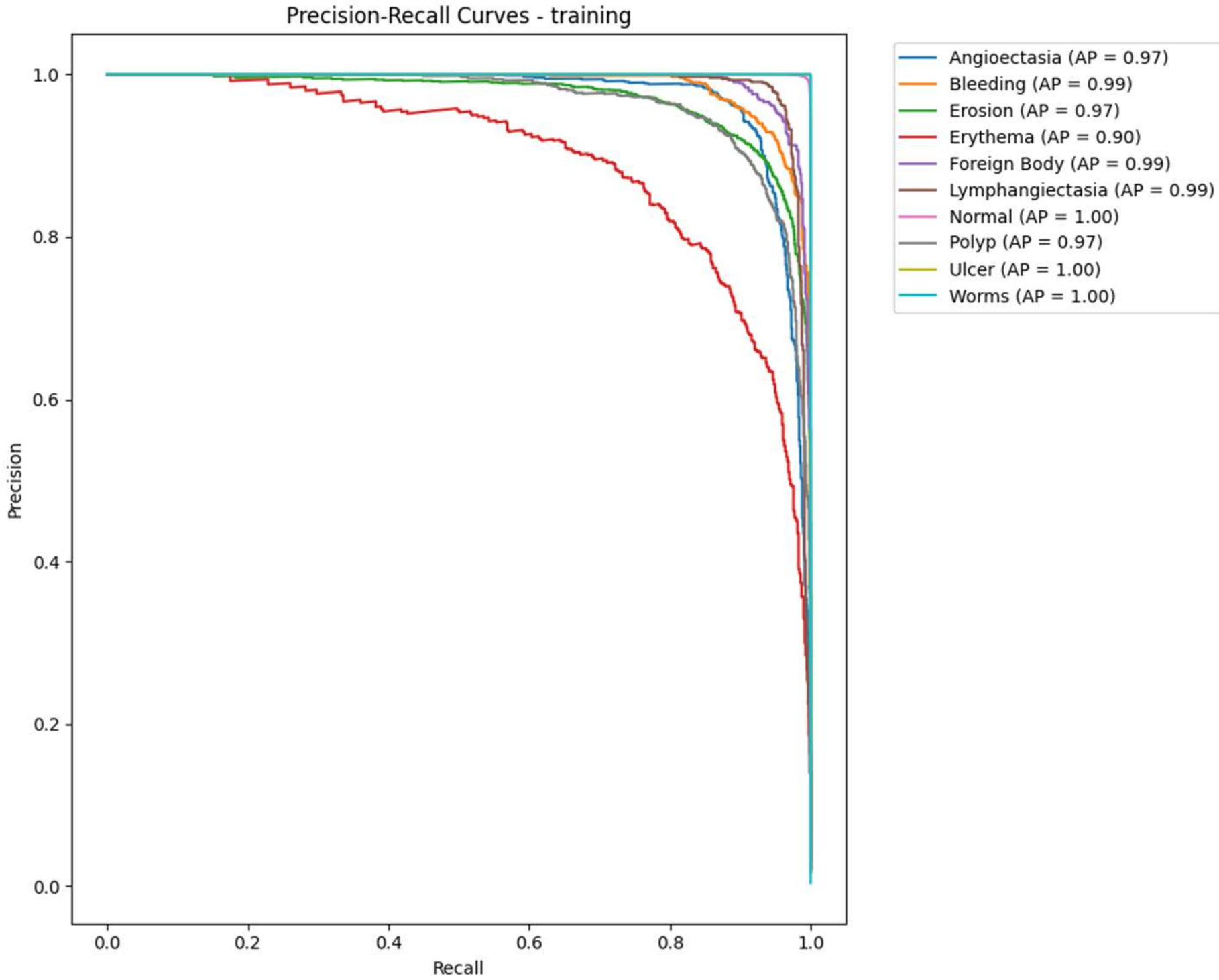

- **Validation Precision-Recall Curve**:

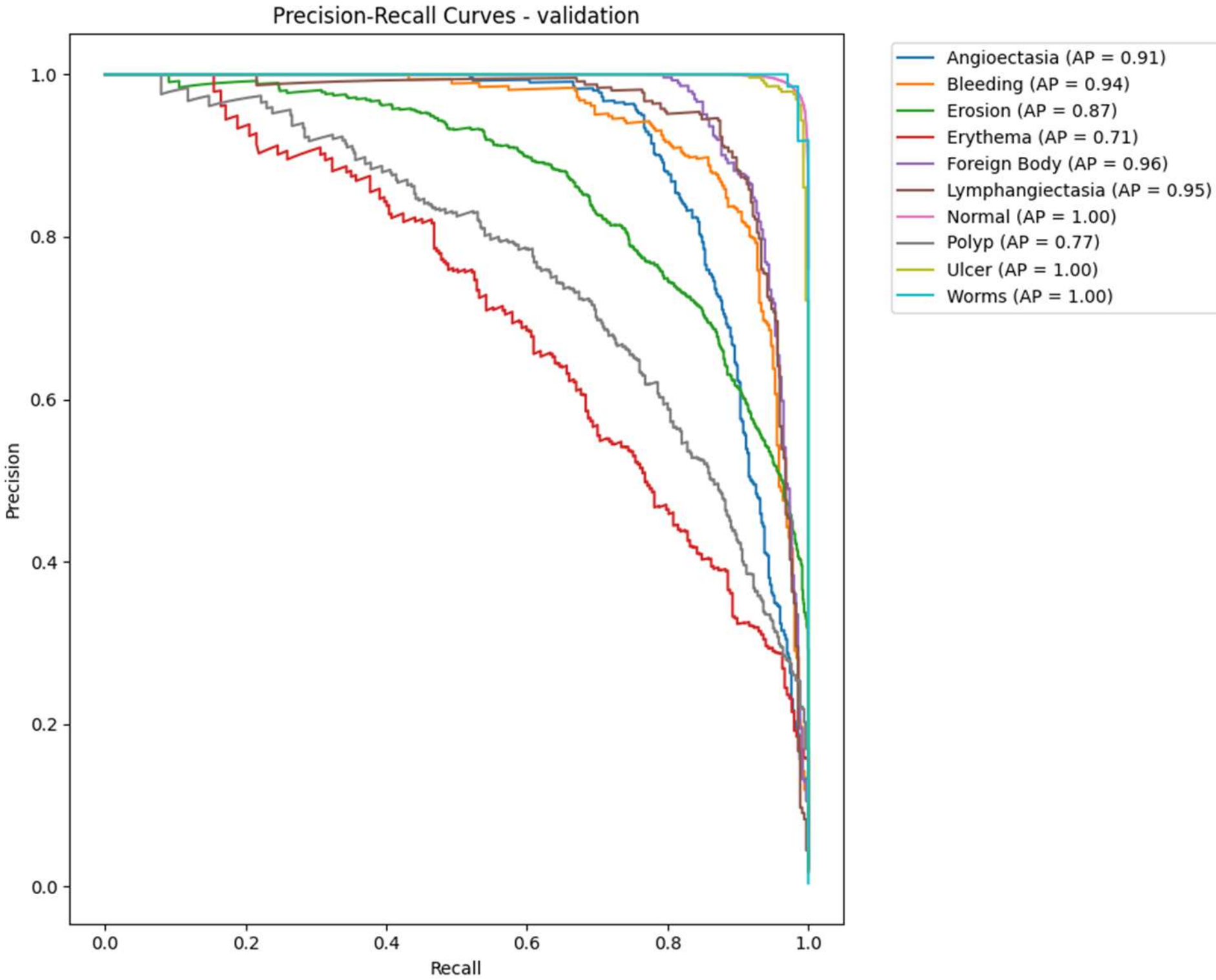

- This curve shows a good balance for most classes, though some challenging classes may need further threshold tuning to improve recall without sacrificing precision.
- The **precision-recall curves** visualize the trade-off between precision (the proportion of correctly predicted positive instances out of all instances predicted as positive) and recall (the proportion of correctly predicted positive instances out of all actual positive instances) across different thresholds.
- The **training precision-recall curves** illustrate high average **precision (AP) values for most classes, with values ranging from 0.97 for Angioectasia to 1.00 for Ulcer and Worms.**
- The **validation precision-recall curves** show slightly lower AP values, particularly for **"Erythema" (0.71), "Foreign Body" (0.96), and "Polyp" (0.77).**
- These curves suggest that while the model generally achieves good performance, some classes, such as **"Erythema" and "Polyp",** may benefit from further threshold tuning to improve recall without significantly sacrificing precision

## 6 Discussion

The BiomedCLIP-PubMedBERT model demonstrated strong performance in fine-tuning for multi-class classification of abnormalities in Video Capsule Endoscopy (VCE) images. This section discusses the model's key strengths, limitations, and potential clinical implications.

### 6.1 Key Findings

The performance analysis provides valuable insights:

- **Classification Challenges:** Varying AUC-ROC scores across abnormality types indicate the challenge of handling class imbalance. **High specificity (0.90)** suggests effective identification of negative cases, but the model faced challenges in sensitivity, particularly in detecting positive instances for some classes.

- **Performance Stability:** Consistent specificity and stable **balanced accuracy around 0.94** suggest that while the model reliably identifies negatives, it may benefit from further optimization to improve overall classification accuracy.

### 6.2 Clinical Implications

This model could significantly impact diagnostic efficiency in gastrointestinal health:

- **Increased Diagnostic Speed:** Automating the classification of VCE frames allows clinicians to focus on images flagged as abnormal, improving diagnostic efficiency.
- **Reduced Human Error:** The AI model's consistent pattern recognition reduces the likelihood of oversight, a common issue in manual review.
- **Vendor Independence:** The model's adaptability to different datasets and imaging conditions supports integration across clinical workflows.

### 6.3 Limitations and Challenges

Despite promising results, some challenges remain:

- **Data Imbalance:** Underrepresented classes, such as "Lymphangiectasia" and "Foreign Body," affected model performance. Although augmentation helped, limited diversity in these classes restricted the model's generalization ability.
- **Visual Similarities Between Classes:** Visually similar classes, such as "Erosion" and "Erythema," challenged the model, underscoring the complexities of medical image classification.
- **Dependence on Pre-trained Text Embeddings:** PubMedBERT embeddings, while domain-specific, may lack nuanced clinical context for rarer conditions. Future advancements in medical language models could address this limitation.

### 6.4 Future Directions

- **Incorporating Temporal Features:** Given VCE's sequential nature, integrating temporal features (e.g., via RNNs or spatio-temporal transformers) could enhance the model's ability to differentiate similar abnormalities.
- **Active Learning:** Implementing active learning to annotate uncertain cases would improve low-confidence regions, minimizing the need for extensive manual labeling.
- **Interpretability Enhancements:** Exploring interpretable AI methods, such as attention-based mechanisms or visual heatmaps, could enhance transparency and clinician trust in the model's outputs.
- **Expansion to Other Modalities:** This model shows potential for adaptation to other medical imaging domains (e.g., CT, MRI, X-ray), widening its applicability with minimal retraining.
- **Model Interpretability:** Exploring interpretable AI methods, such as attention-based mechanisms or generating visual heatmaps, could help clinicians understand how the model arrived at a specific diagnosis. This would enhance trust and transparency in clinical settings.
- **Cross Attention Mechanism:** We would like to do Cross Attention Mechanism during Similarity matching to get more accuracy and Attention Mechanis make the model to remember the learned knowledge for longer dependencies.

### 6.3 Ethical and Regulatory Considerations

When applying AI models in medical diagnostics, it is essential to consider the ethical and regulatory implications. Any clinical application of the model would need to undergo thorough validation and regulatory approval to ensure its safety and effectiveness. Additionally, measures must be taken to ensure that patient data used in training and evaluation is handled securely and complies with relevant privacy regulations (e.g., HIPAA).

In clinical practice, AI models should function as decision-support systems rather than replacing human expertise. The goal is to enhance clinician productivity and reduce diagnostic errors, while ensuring that the final decisions remain in the hands of trained medical professionals.

# 7 Conclusion

The goal of this research was to develop an automated, Fine-Tuned classification model for detecting abnormalities in Video Capsule Endoscopy (VCE) frames using the BiomedCLIP-PubMedBERT framework. The study aimed to address the challenges of manual image interpretation in gastrointestinal diagnostics, which is both time-consuming and prone to error. By leveraging pre-trained vision and language models, we successfully created a system capable of classifying ten distinct abnormality categories: Angioectasia, Bleeding, Erosion, Erythema, Foreign Body, Lymphangiectasia, Polyp, Ulcer, Worms, and Normal.

## 7.3 Key Contributions

The primary contributions of this study are:

- **Fine - Tuning Application:** We demonstrated the effective use of BiomedCLIP-PubMedBERT for Fine tuning in medical image classification. By aligning image features with text embeddings, the model could classify abnormalities without needing large annotated datasets.

- **High Accuracy and Generalization:** The model achieved high classification accuracy, precision, recall, and F1 scores across most abnormality classes. Its ability to generalize across unseen validation images is especially important in clinical settings where labeled data may be limited.

- **Integration of Vision and Language Models:** The integration of a Vision Transformer (ViT) with PubMedBERT allowed the model to capture both visual and semantic information, resulting in improved classification performance. The model's success in mapping images to text-based class labels demonstrates the potential of multimodal models for medical diagnostics.

- **Scalability and Vendor Independence:** The model is scalable and adaptable to various clinical settings, thanks to its pre-trained nature and fine-tuning capabilities. It can work across different datasets and equipment without requiring extensive retraining.

## 7.4 Final Remarks

The development of AI-based models for medical image classification is a rapidly growing field, with the potential to revolutionize diagnostic workflows. This study contributes to that effort by introducing a novel fine-tuning approach for abnormality classification in endoscopic images. By reducing the need for extensive manual labeling, our model holds promise for improving diagnostic efficiency, reducing human error, and ensuring more timely and accurate medical diagnoses.

As the healthcare field continues to embrace AI-driven tools, models like BiomedCLIP-PubMedBERT will play an increasingly important role in enhancing the capabilities of clinicians and improving patient outcomes. With further refinement and validation, such models could become an integral part of future medical imaging systems.

# 8 Acknowledgments

As participants in the Capsule Vision 2024 Challenge, we fully comply with the competition's rules as outlined in [1]. Our AI model development is based exclusively on the datasets provided in the official release in [2].

We also acknowledge the developers of the BiomedCLIP and PubMedBERT models for their contributions to the open-source community, enabling us to leverage these state-of-the-art tools for medical image classification.

Additionally, we would like to thank our colleagues and collaborators who provided insightful feedback during the development and evaluation stages of the project. Their input was invaluable in refining the approach and improving the overall performance of the model.